\documentclass[letterpaper, 10 pt, journal, twoside]{IEEEtran}
\usepackage[left=16.9mm,right=16.9mm,top=20.1mm,bottom=15.2mm]{geometry}
\usepackage[utf8]{inputenc}
\usepackage[T1]{fontenc}

\usepackage{amsmath,amsfonts}
\usepackage{array}
\usepackage{graphicx}
\usepackage[caption=false,font=small,labelfont=sf,textfont=sf]{subfig}
\usepackage{textcomp}
\usepackage{stfloats}
\usepackage{url}
\usepackage{cite}
\usepackage{xcolor}
\usepackage{multirow}
\usepackage{booktabs}
\usepackage{siunitx}
\usepackage{hyperref}
\usepackage{acro}

\graphicspath{{figures/}}
\newcommand{\newac}[2]{\DeclareAcronym{#1}{short=#1,long=#2}}
\newac{DEM}{Digital Elevation Map}
\newac{GNC}{Guidance{,} Navigation{,} and Control}
\newac{GPU}{Graphics Processing Unit}
\newac{IMU}{Inertial Measurement Unit}
\newac{IoU}{Intersection over Union}
\newac{LRO}{Lunar Reconnaissance Orbiter}
\newac{MCP}{Model Context Protocol}
\newac{SPAD}{Single-Photon Avalanche Diode}
\newac{FMM}{Fast Marching Method}
\newac{DWA}{Dynamic Window Approach}

\begin{document}

\title{MARE: An Open-Source Lunar Rover Simulator with Shadow Ground Truth for Perception and Autonomy}

\author{R.~Castilla-Arquillo$^{1}$ and M.~Olivares-Mendez$^{1}$
\thanks{$^{1}$R. Castilla-Arquillo and M. Olivares-Mendez are with the Space Robotics (SpaceR) Research Group, SnT, University of Luxembourg, Luxembourg (e-mail: raul.castilla@uni.lu).}}

\maketitle

\begin{abstract}
Lunar surface exploration increasingly relies on autonomous rovers that must perceive and traverse unstructured terrain under extreme illumination conditions. Near the lunar poles, low sun elevations produce long, high-contrast shadows that challenge stereo vision, terrain segmentation, and localization. Developing perception and navigation algorithms for these conditions requires large amounts of sensor data with accurate ground truth, which field campaigns and analog facilities can only provide at high cost and with limited variety. This paper presents MARE (Modular Autonomy and Rendering Environment), an open-source simulation suite for lunar rover perception and autonomy built on Unreal Engine~5 and ROS~2. MARE consists of three integrated components. Terrain Studio generates cratered and rocky lunar Digital Elevation Maps (DEMs) to build simulation scenes. A dedicated Unreal Engine project hosts two physics-based rovers with plugins for stereo, lidar, event, and Single-Photon Avalanche Diode (SPAD) cameras, streaming time-stamped sensor data and pixel-accurate ground truth. Lastly, a Guidance, Navigation, and Control (GNC) stack provides web-based mission control. Validation shows a mean Intersection over Union (IoU) of 0.79 between predicted and rendered shadows at low sun elevations, and stereo depth analysis verifies decreased pixel coverage in shadowed regions relative to sunlit terrain. A dataset of 5040 frames recorded with MARE is released, on which segmentation baselines reach a mean IoU of 0.85 for shadows and 0.60 for objects on unseen terrain.
\end{abstract}

\section{INTRODUCTION}
\label{sec:introduction}

\suppressfloats[t]
\begin{figure}[t]
  \centering
  \includegraphics[width=\columnwidth]{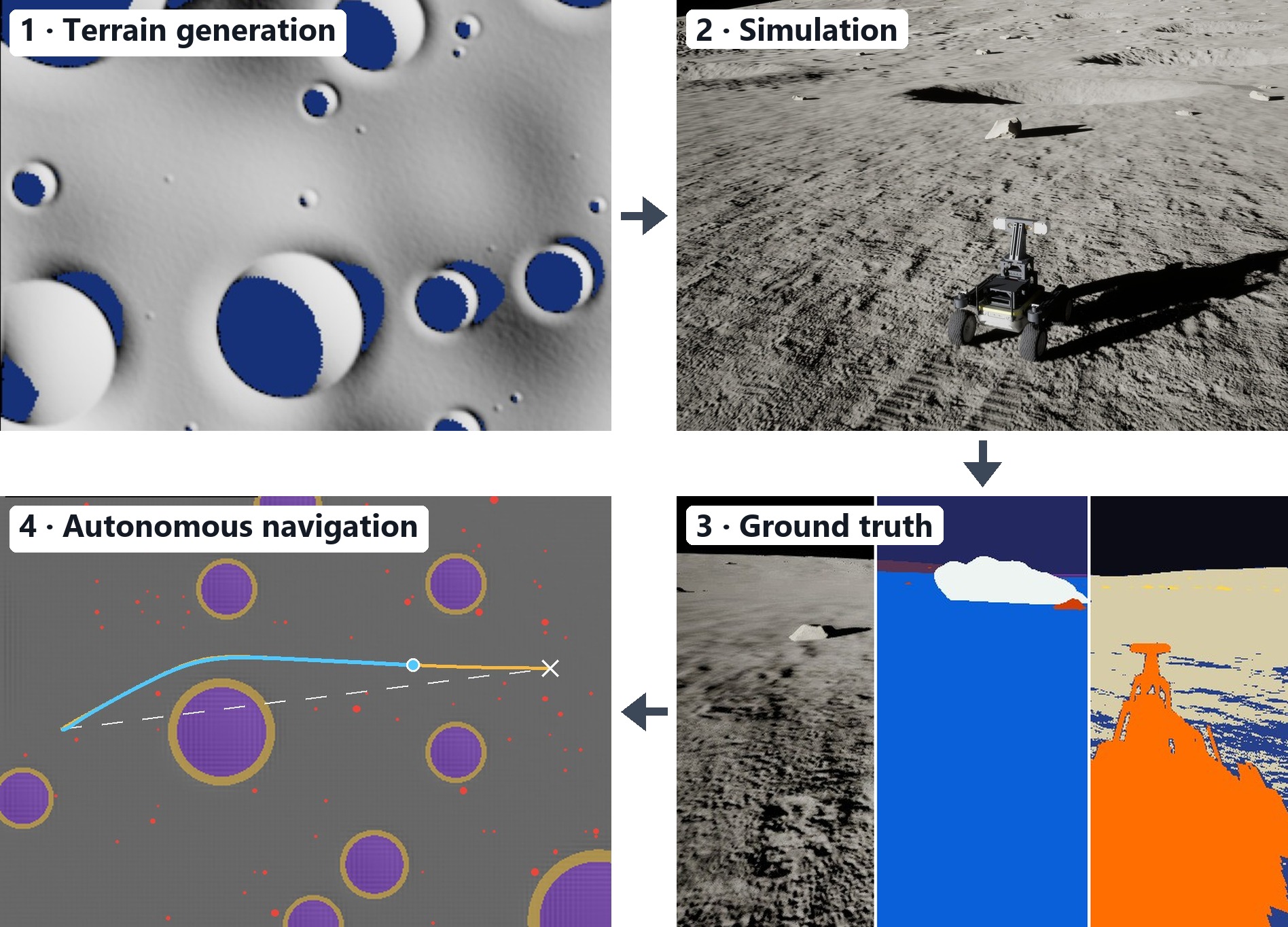}
  \caption{\small Overview of MARE: (1) terrain generation, (2) simulation, (3) ground truth, and (4) autonomous navigation.}
  \label{fig:overview}
\end{figure}

\begin{figure*}[t]
  \centering
  \includegraphics[width=\textwidth]{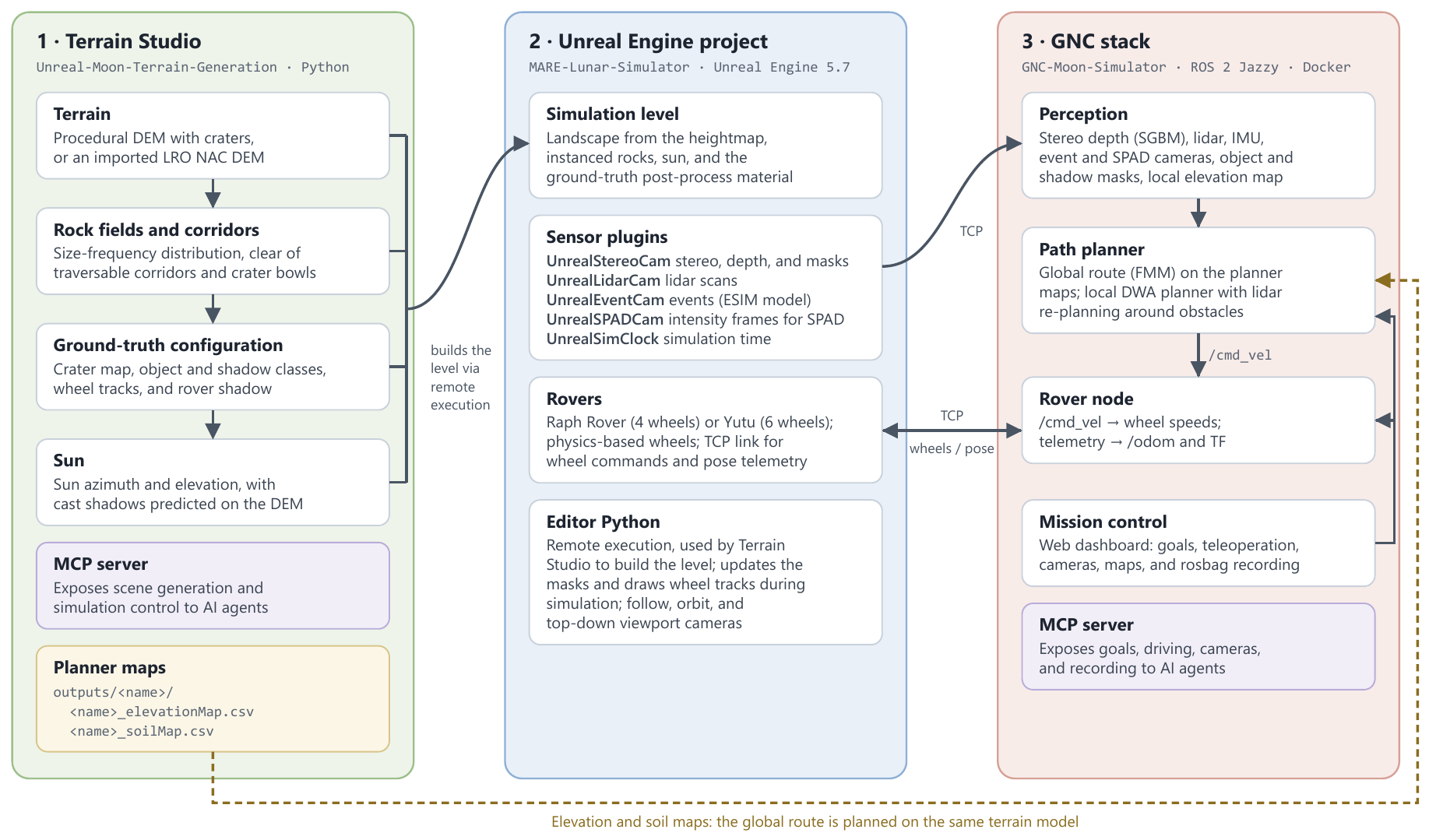}
  \caption{\small Architecture of the MARE simulation suite.}
  \label{fig:pipeline}
\end{figure*}

Planetary rovers depend on their \ac{GNC} system to traverse complex terrain autonomously, since communication delays limit direct human control. On the Moon, extreme illumination conditions complicate this task: near the poles, the sun remains only a few degrees above the horizon, casting long, high-contrast shadows that continuously alter the terrain's appearance over time~\cite{hansen2024polar, rodriguez2026spice}. Developing and validating perception and navigation algorithms for these environments requires large volumes of sensor data paired with accurate ground truth, including object labels, shadow masks, depth maps, and the rover pose.

While field campaigns at analog sites and laboratories produce realistic datasets, they are costly and provide little control over terrain geometry and illumination. Simulation bridges this gap by generating labeled data within strictly controlled, repeatable environments. Building on the success of terrestrial game-engine simulators like CARLA~\cite{dosovitskiy2017carla} and AirSim~\cite{shah2017airsim}, recent work has adapted high-fidelity rendering for planetary contexts. Allan \emph{et al.}~\cite{allan2019planetary} developed a Gazebo simulator utilizing physically based rendering and fractal terrain for polar rovers. OmniLRS~\cite{richard2024omnilrs} leverages NVIDIA Isaac Sim to offer ray-traced lunar rendering, terrain deformation, and mission integration. Similarly, Space Robotics Bench~\cite{orsula2025space} applies procedural generation and domain randomization in parallel environments for zero-shot sim-to-real learning. Other notable platforms include LunarSim~\cite{pieczynski2023lunarsim}, which pairs Unity's visuals with ROS~2 for computer vision evaluation, and MarsSim~\cite{zhou2022marssim}, which links multiscale Martian landscapes with a Gazebo wheel-soil terramechanics plugin.

Despite these advances, planetary simulation tools remain fragmented. A recent survey~\cite{kimsimulation} noted that only eight of fifteen reviewed simulators are publicly available, limiting reproducibility in the community. Current systems typically isolate specific domains like rendering, policy learning, or vehicle dynamics, rather than providing a unified pipeline from terrain creation to closed-loop autonomy with pixel-accurate labeling. Bridging this gap requires complete stack consistency: the same terrain model used to build the scene must also generate the ground truth and serve as the rover's planning map, ensuring labels, maps, and sensors all operate within the same world.

Within this ground truth, shadows and wheel tracks deserve particular attention. Because the Moon lacks an atmosphere, shadowed regions remain almost completely dark; a shadow can conceal a rock or crater, eliminate the surface texture needed for stereo matching, or be mistaken for an obstacle. Consequently, rock and crater detection depends strongly on illumination~\cite{castro2024mask}. At low sun elevations, the rover's own shadow also extends far ahead of or behind the vehicle, occupying a large portion of the navigation camera images. Because this shadow moves with the rover, it undermines the static-scene assumption of visual odometry and compromises terrain segmentation. Similarly, the wheel tracks the rover leaves alter the appearance of the traversed surface. Perception algorithms operating under these conditions must therefore be trained and evaluated with labels that distinguish sunlit from shadowed terrain, the rover's self-shadow from static terrain shadows, and wheel tracks from the undisturbed surface, which existing planetary simulators rarely provide.

These illumination extremes also motivate sensing modalities beyond conventional cameras and lidar. Emerging sensors, such as event cameras and \ac{SPAD} cameras, are of growing interest for lunar perception under low- and high-contrast lighting~\cite{rodriguez2026spice}. However, they remain largely absent from existing planetary simulators: none of the surveyed platforms supports event cameras, and only one provides limited \ac{SPAD} support~\cite{kimsimulation}.

To address these gaps, we present MARE (Modular Autonomy and Rendering Environment), an open-source simulation suite for lunar rover perception and autonomy built on Unreal Engine~5 and ROS~2. Figure~\ref{fig:overview} shows its end-to-end pipeline. Terrain Studio generates procedural or real lunar terrain with craters and rock fields and builds the scene in Unreal Engine. Two physics-based rovers carry reusable sensor plugins, including stereo, lidar, event, and \ac{SPAD} cameras, synchronized by a shared simulation clock. A ground-truth module produces pixel-accurate object and shadow masks for craters, crater rims, rocks, wheel tracks, and the rover's own shadow. Finally, a containerized \ac{GNC} stack drives the rover autonomously, supervised from a web-based mission control. We validate the suite in three ways: by comparing shadows predicted from the terrain model with those rendered by the engine on three terrains, by evaluating stereo depth separately on sunlit and shadowed terrain using the ground-truth masks, and by training segmentation and depth networks on a dataset recorded with the suite, which we release. The source code is publicly available on GitHub\footnote{\url{https://github.com/raulcastar/MARE-Lunar-Simulator}}, together with a video overview\footnote{\url{https://youtu.be/EI4HgbMFbBE}}.

\section{System Overview}
\label{sec:overview}

MARE runs on a single workstation and consists of three components, each released as an open-source repository. As shown in Figure~\ref{fig:pipeline}, Terrain Studio and the Unreal Engine editor run on Windows, while the \ac{GNC} stack runs under WSL2 as one Docker container per ROS~2 \cite{macenski2022robot} node. This section outlines each component and how they communicate.

MARE is built on Unreal Engine~5. The engine combines real-time global illumination and virtual shadow maps, which reproduce the long, sharp shadows of a low sun, with landscape tools that scale to kilometer-sized terrains. It also runs on consumer-grade graphics hardware, such as the laptop used in this work, supports virtual reality headsets natively for immersive scene inspection, and has a large user community. Unreal Engine provides no robotics interfaces, so we implement ROS~2 communication, sensor models, and time synchronization as reusable plugins, described in Section~\ref{sec:rovers}.

\textbf{Terrain Studio (scene generation):} Generates the scene in four steps: terrain, rock fields and corridors, ground-truth configuration, and sun. Using Python remote execution, it builds the simulation level in the open Unreal Engine editor with a single action. It also exports terrain elevation and soil maps, which the \ac{GNC} stack uses for path planning. The implementation is available on GitHub\footnote{\url{https://github.com/raulcastar/Unreal-Moon-Terrain-Generation}}.

\textbf{Unreal Engine project (simulation and rendering):} Contains the simulation level, sensor plugins, physics-based rovers, and the editor Python scripts Terrain Studio uses to build the level. During simulation, each sensor plugin renders its own captures and streams them over TCP to a dedicated ROS~2 node. The rover exchanges wheel commands and pose telemetry with the GNC stack.

\textbf{GNC stack (autonomy and operations):} Comprises perception, path-planning, and rover nodes. Together, they turn sensor streams into depth and local maps, plan routes on the planner maps, and command the rover. A web-based mission control supervises the operation. The implementation is available on GitHub\footnote{\url{https://github.com/raulcastar/GNC-Moon-Simulator}}.

All sensor data is time-stamped with the Unreal Engine simulation time, which the UnrealSimClock plugin also provides to the ROS~2 nodes as their clock. As a result, the autonomy stack stays consistent with the simulation even when rendering slows down, or the simulation is paused, and recorded data replays with the timestamps observed at capture.

\section{Terrain and Scene Generation}
\label{sec:terrain}

\begin{figure*}[t]
  \centering
  \includegraphics[width=\textwidth]{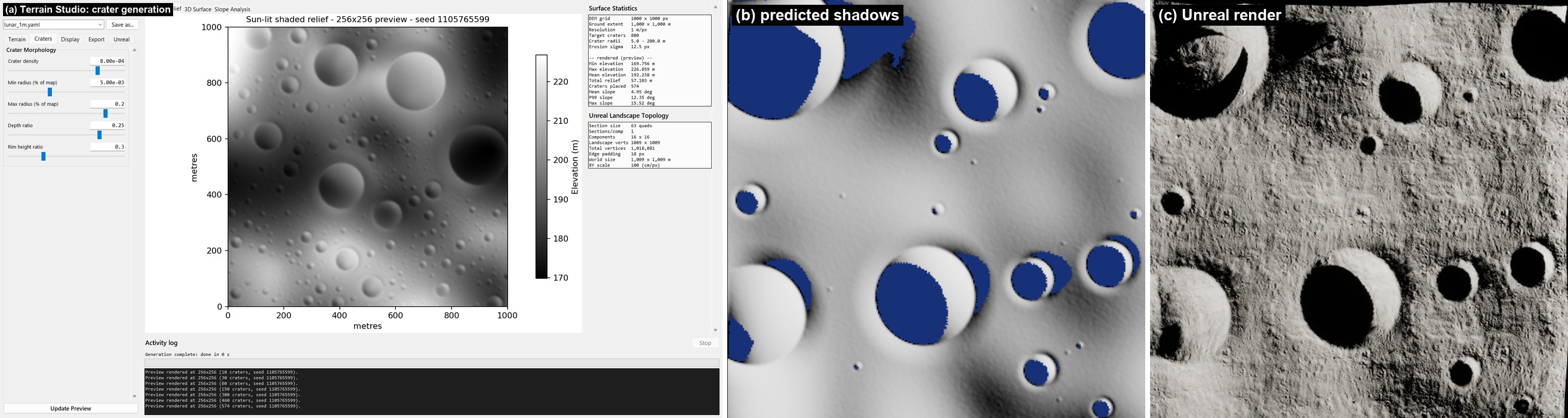}
  \caption{\small Terrain Studio: (a) terrain generation, (b) predicted cast shadows, and (c) rendered terrain.}
  \label{fig:terrain}
\end{figure*}

Terrain Studio either generates a synthetic \ac{DEM} or loads a real one. A \ac{DEM} is a 2D map of elevation values, and it can have different resolutions. Terrain Studio computes each step on this \ac{DEM} and then transfers the result to Unreal Engine, so the scene, the ground truth, and the planner share the same terrain model. This section covers the terrain, the rock fields and corridors, and the sun. Section~\ref{sec:rovers} covers the ground-truth configuration.

\textbf{Terrain:} Synthetic \acp{DEM} combine smoothed macro-relief, micro-roughness, and non-overlapping craters. Crater depth and rim height are set as configurable ratios of the crater radius. A live preview shows the result (Figure~\ref{fig:terrain}a). Real \acp{DEM}, such as \ac{LRO} Narrow Angle Camera products \cite{robinson2010lunar}, can be loaded instead. The generator records every crater it places, and this list later serves as exact ground truth.

\textbf{Rock fields and corridors:} Rocks follow a size-frequency distribution scaled to the rock abundance measured by the \ac{LRO} Diviner instrument \cite{bandfield2011lunar}. To provide traversable terrain for autonomy experiments, Terrain Studio generates clear driving corridors. It identifies impassable craters and calculates pathways using a Voronoi diagram of their centers, ensuring that both the corridors and crater bowls remain free of rocks.

\textbf{Sun:} Terrain Studio defines the sun's azimuth and elevation to predict cast shadows on the \ac{DEM} using the method proposed by Corripio~\cite{corripio2003vectorial}. This technique traverses each \ac{DEM} line along the sun's azimuth toward the light, calculating the shadow height cast by the preceding terrain. Given a cell size $\Delta$ and a sun elevation $\varepsilon$, the shadow height over the $k$-th cell is
\begin{equation}
  s_k = \max\left(s_{k-1},\, z_{k-1}\right) - \Delta \tan\varepsilon,
  \label{eq:shadow1}
\end{equation}
where $z_k$ represents the cell elevation. A cell is considered shadowed if $s_k > z_k$. As illustrated in Figure~\ref{fig:terrain}b, the predicted shadows for a sun elevation of \ang{11} appear in blue. This prediction lets the system prepare a scene for specific lighting conditions before rendering. Finally, Terrain Studio constructs the simulation level within the Unreal Engine editor by generating the landscape, placing rocks, and configuring the sun (Figure~\ref{fig:terrain}c).

\section{Rovers, Sensors, and Ground Truth}
\label{sec:rovers}

Experiments need a vehicle to carry the sensors, sensors to observe the scene, and a reference to judge their output against. In MARE, all three are tied to the same rendered frame. The rovers drive through the generated terrain, the sensor plugins capture what they see, and the ground truth, meaning the exact labels the simulator knows about the scene, is produced alongside the images. This lets perception algorithms be trained and evaluated against what is actually in the scene.

\begin{figure}[t]
  \centering
  \includegraphics[width=\columnwidth]{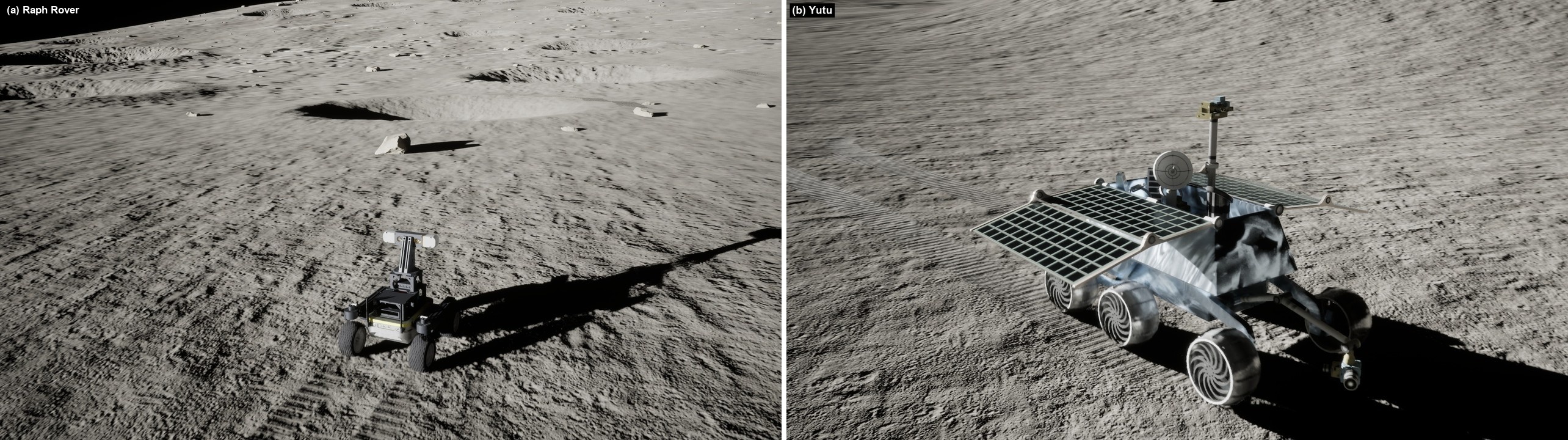}
  \caption{\small Simulated rovers: (a) Raph Rover and (b) Yutu.}
  \label{fig:rovers}
\end{figure}

\textbf{Rovers:} Two physics-based rovers are available, as shown in Figure~\ref{fig:rovers}. The first is a model of the Raph Rover by Fictionlab\footnote{\url{https://fictionlab.pl/raph-rover/}}, driven in simulation as a four-wheeled skid-steer platform. The second is Yutu, a six-wheeled rover with solar arrays and a sensor mast, adapted from a public model. Both rovers share the same Unreal Engine Blueprint logic and communication interface. Switching between them is therefore a configuration change that updates the drive calibration, the path planner parameters, and the robot description without restarting the stack. Both rovers also carry dimmable headlights, controlled from the \ac{GNC} stack, which illuminate the terrain ahead of the rover when no sunlight reaches it, as in permanently shadowed regions.

\begin{figure}[t]
  \centering
  \includegraphics[width=\columnwidth]{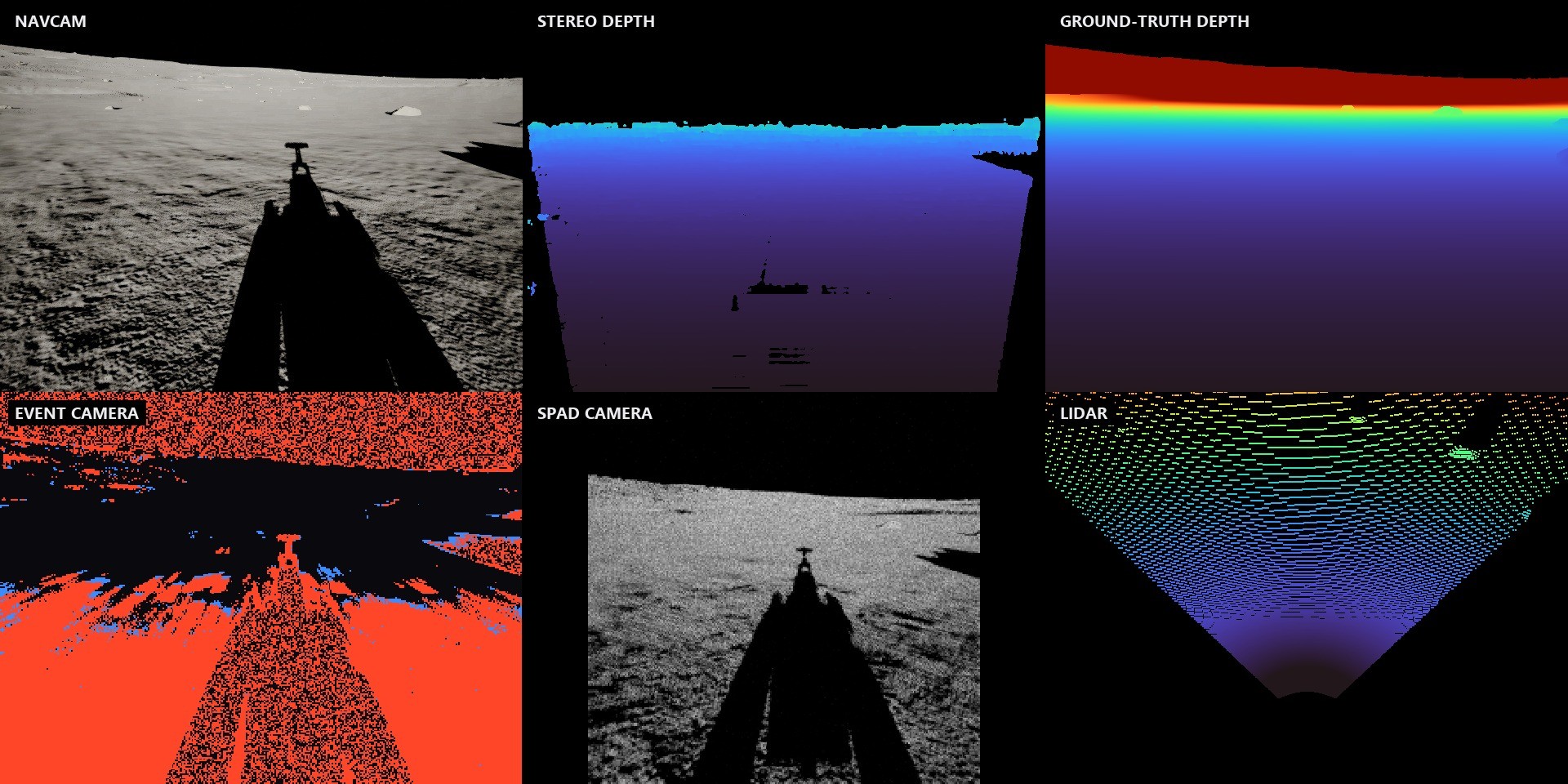}
  \caption{\small Simulated sensor outputs.}
  \label{fig:sensors}
\end{figure}

\textbf{Sensor plugins:} Each sensor is an independent Unreal Engine plugin. Figure~\ref{fig:sensors} shows synchronized outputs of the navigation camera, stereo depth, ground-truth depth, event camera, \ac{SPAD} camera, and lidar. UnrealStereoCam provides a stereo rig of navigation cameras with a rear camera, ground-truth depth, and object masks. UnrealLidarCam provides a lidar with a configurable aperture and beam pattern. UnrealEventCam provides an event camera based on the ESIM model \cite{rebecq2018esim}. UnrealSPADCam provides a camera whose intensity frames feed a \ac{SPAD} photon model ported from spadSim \cite{spadsim}. UnrealSimClock shares the simulation time among all plugins, and an \ac{IMU} is simulated from the rover odometry. The plugins do not depend on ROS and can be reused in other Unreal Engine projects; each is also released as a standalone repository.

\begin{figure}[t]
  \centering
  \includegraphics[width=\columnwidth]{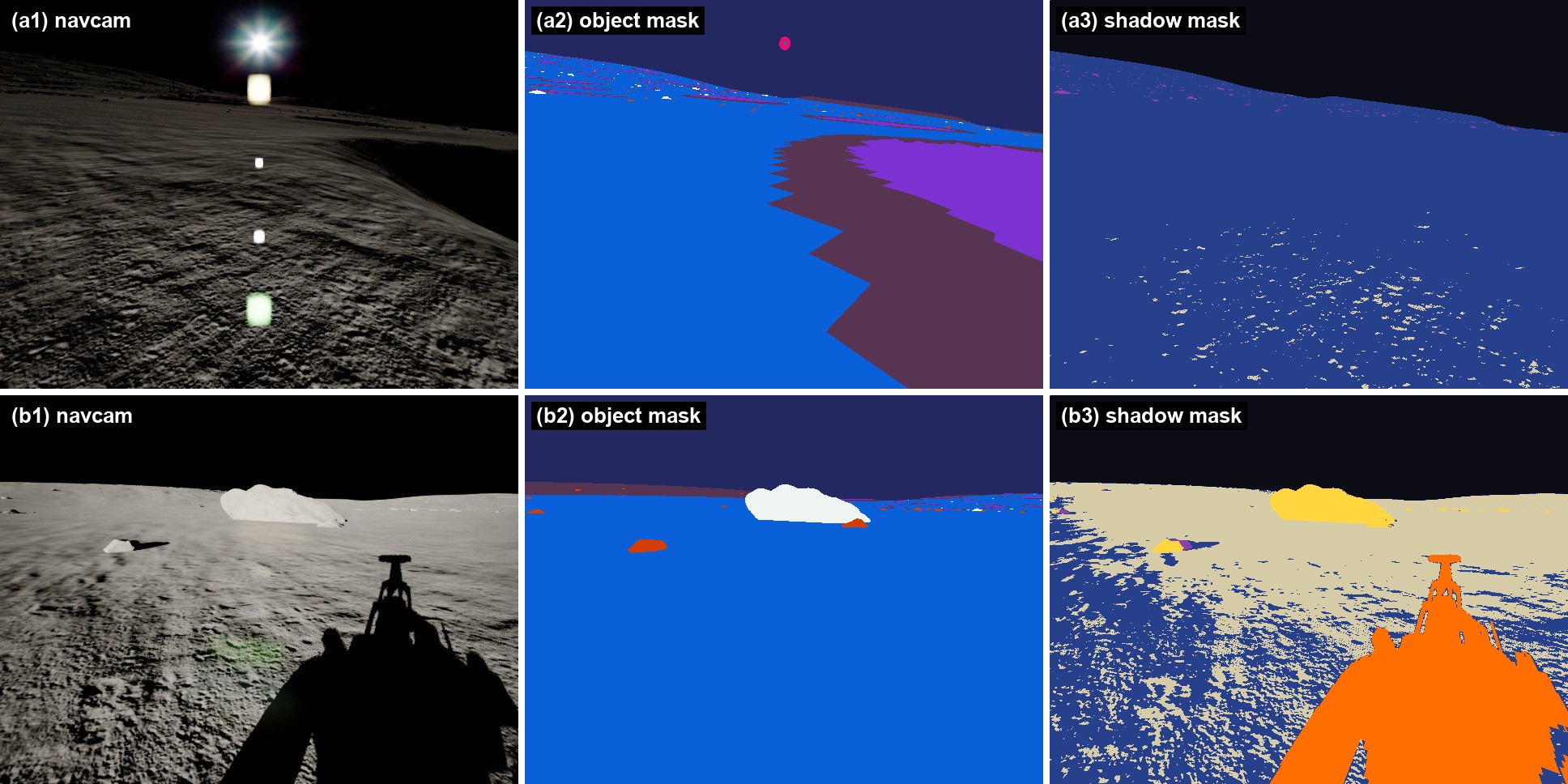}
  \caption{\small Ground-truth masks: (a) along a crater rim and (b) approaching a boulder.}
  \label{fig:gt}
\end{figure}

\textbf{Ground truth:} A post-processing material on a dedicated mask camera assigns two labels to every pixel, as illustrated in Figure~\ref{fig:gt}:
\begin{itemize}
  \item \textbf{Object class:} sky, sun, landscape, crater, crater rim, wheel track, boulder, small rock, and rover.
  \item \textbf{Shadow class:} sunlit and shadowed ground, sunlit and shadowed objects, and the rover's own shadow.
\end{itemize}
The labels are produced in three ways:
\begin{itemize}
  \item Rocks and the rover use custom stencil values assigned at placement.
  \item Craters and rims are projected onto the landscape from a crater map derived from the generator's crater list.
  \item The rover's shadow is separated by rendering the rover's depth from the sun.
\end{itemize}

Each mask is streamed synchronously with its corresponding rendered image frame. As illustrated in Figure~\ref{fig:gt}a, the object mask delineates the crater floor and rim. Figure~\ref{fig:gt}b shows an object mask identifying boulders and small rocks, alongside a shadow mask that distinguishes between sunlit rock, shadowed rock, and the rover's own cast shadow. These specific labels target the visual challenges outlined in Section~\ref{sec:introduction}. Consequently, the general shadow class enables isolated evaluation of perception algorithms on sunlit versus shadowed terrain, the rover-shadow class isolates the dynamic shadow moving with the vehicle, and the wheel-track class highlights terrain altered by the rover's traversal.

\section{Autonomy and Operations}
\label{sec:autonomy}

\begin{figure}[t]
  \centering
  \includegraphics[width=\columnwidth]{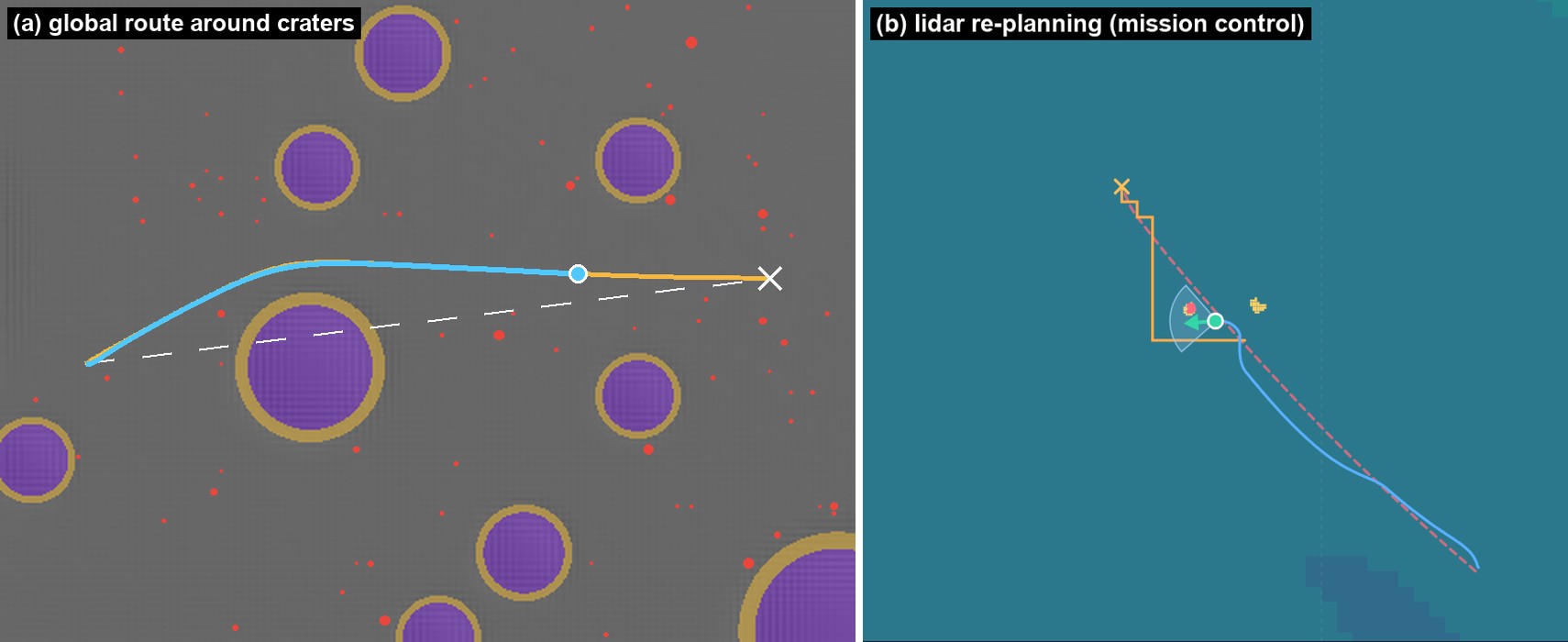}
  \caption{\small Autonomous navigation: (a) global route around a crater and (b) lidar-based re-planning.}
  \label{fig:planning}
\end{figure}

Autonomous navigation through a simulated environment requires a robust \ac{GNC} stack, complemented by interfaces that let both human operators and AI agents monitor and command the vehicle. This section details these operational components, with Figure~\ref{fig:planning} illustrating two examples of such autonomous drives in action.

\textbf{Perception:} The perception module computes stereo depth with semi-global block matching (SGBM)~\cite{hirschmuller2008stereo} and builds a local elevation map around the rover. It publishes the lidar, \ac{IMU}, event-camera, and \ac{SPAD} data as ROS~2 topics, together with the object and shadow masks.

\textbf{Path planner:} Global trajectories come from a \ac{FMM} planner~\cite{sanchez2022optimal} based on DyMu~\cite{dymu}. It runs on the maps exported by Terrain Studio, which are the same \ac{DEM} and soil map used to build the scene, and treats slopes above \ang{15} as impassable. A \ac{DWA} local planner~\cite{fox1997dynamic} follows the route, and re-planning starts whenever the lidar detects an obstacle blocking it. Figure~\ref{fig:planning}a demonstrates this: the direct dashed line crosses a steep crater, forcing the global (gold) and driven (blue) paths to circumvent it. Figure~\ref{fig:planning}b further illustrates the system re-planning around unmapped rocks omitted from the coarse global map. To execute these movements, a rover node converts the resulting velocity commands into wheel speeds and provides odometry feedback.

\begin{figure*}[t]
  \centering
  \includegraphics[width=\textwidth]{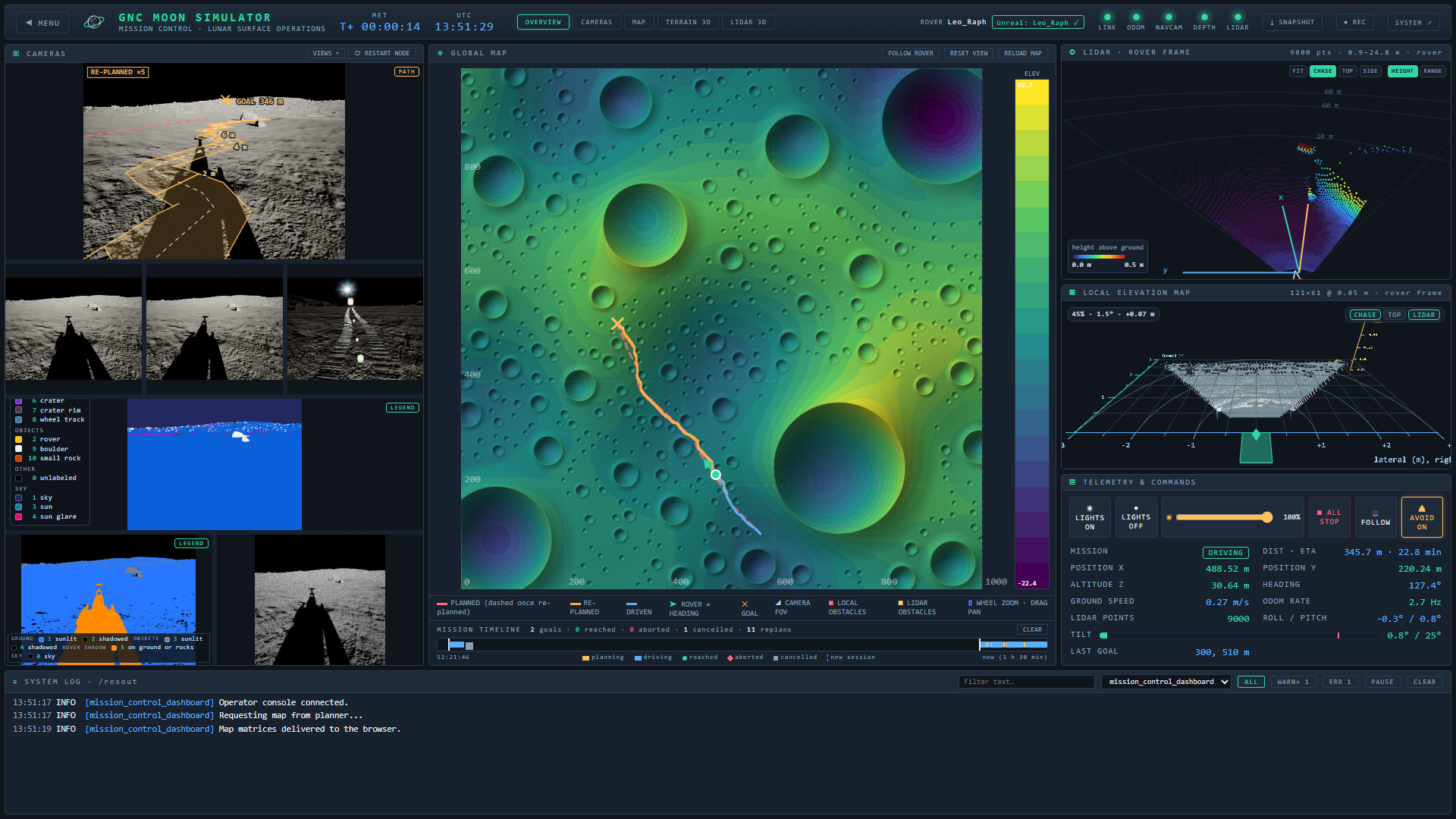}
  \caption{\small Web-based mission control during an autonomous drive. The global map shows the whole terrain, with relief shading that makes its craters visible, and a planned route of about \SI{500}{m} between the large craters.}
  \label{fig:mission}
\end{figure*}

\textbf{Mission control:} The web-based dashboard, illustrated in Figure~\ref{fig:mission}, features a three-column interface. The left column displays the navigation camera feeds alongside the planned path and ground-truth masks. The central column presents the global map, detailing the initial, re-planned, and actual driven routes. The right column visualizes the lidar data, local elevation map, and rover telemetry, all situated above the mission timeline and system log. Through this interface, operators can assign navigation goals, teleoperate the vehicle, and manage rosbag recordings and playbacks.

\textbf{MCP servers:} Terrain Studio and the \ac{GNC} stack each feature a \ac{MCP} server that makes their operations accessible to AI agents. Through these interfaces, agents can procedurally generate terrain, set up the scene, launch the simulation, and command the rover.

\section{Experiments}
\label{sec:experiments}

We validated MARE through four evaluations: comparing shadow predictions from Terrain Studio with the final Unreal Engine renders, assessing stereo depth accuracy across sunlit and shadowed terrain, training perception baselines on a recorded dataset, and benchmarking the suite's overall runtime. All tests were conducted on a laptop equipped with an NVIDIA GeForce RTX~4060 Laptop GPU, running Unreal Engine~5.7.4 and ROS~2 Jazzy.

\subsection{Shadow Prediction}

\begin{figure}[t]
  \centering
  \includegraphics[width=\columnwidth]{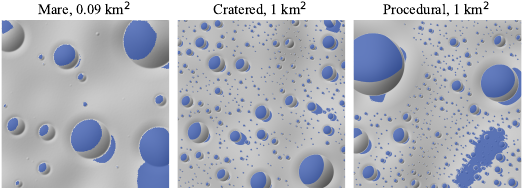}
  \caption{\small Terrains of the shadow evaluation, with the predicted shadows (blue) for a sun at \ang{10} elevation.}
  \label{fig:terrains}
\end{figure}

\begin{table}[t]
\centering
\setlength{\tabcolsep}{4pt}
\renewcommand{\arraystretch}{1.2}
\footnotesize
\caption{\small Terrains of the shadow evaluation.}
\label{tab:terrains}
\begin{tabular}{lccccc}
\toprule
Terrain & Area & Craters (radius) & Relief & Mean slope & \acs{IoU} \\
\midrule
Mare       & \SI{0.09}{km^2} & 26 (\SIrange{2}{49}{m})   & \SI{26}{m}  & \ang{10.4} & 0.78 \\
Cratered   & \SI{1}{km^2}    & 400 (\SIrange{5}{103}{m}) & \SI{66}{m}  & \ang{9.1}  & 0.79 \\
Procedural & \SI{1}{km^2}    & 400 (\SIrange{5}{186}{m}) & \SI{111}{m} & \ang{10.8} & 0.78 \\
\bottomrule
\end{tabular}
\\[2pt]
\parbox{\columnwidth}{\scriptsize Relief: difference between the highest and lowest elevation. \acs{IoU}: mean over sun elevations of \ang{25} and below.}
\end{table}

We compared the cast shadows predicted by Terrain Studio with those rendered by Unreal Engine on three terrains, all generated with Terrain Studio and shown in Figure~\ref{fig:terrains} and Table~\ref{tab:terrains}: \emph{Mare}, a \SI{0.09}{km^2} plain with few craters; \emph{Cratered}, a \SI{1}{km^2} field of small and medium craters; and \emph{Procedural}, a \SI{1}{km^2} terrain with larger craters and higher relief. Each terrain was evaluated at ten sun elevations between \ang{3} and \ang{35} and at two azimuths, \ang{120} and \ang{300}, for 60 configurations in total.

For each configuration, an orthographic top-down camera aligned with the \ac{DEM} footprint captured the terrain in two passes: one with cast shadows and one without. A cell was classified as shadowed if its luminance in the shadowed render dropped below half of its unshadowed luminance, or if it remained dark in both renders (such as on slopes oriented away from the sun). Because this classification relies on a differential comparison between the two renders, the resulting shadow labels are invariant to local albedo and slope shading.

\begin{figure}[t]
  \centering
  \includegraphics[width=\columnwidth]{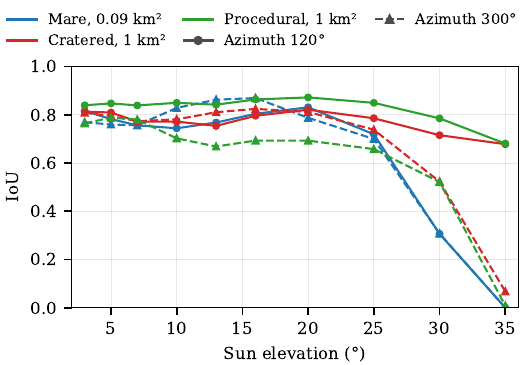}
  \caption{\small \acs{IoU} between predicted and rendered shadows as a function of sun elevation.}
  \label{fig:shadow-iou}
\end{figure}

\begin{figure}[t]
  \centering
  \includegraphics[width=\columnwidth]{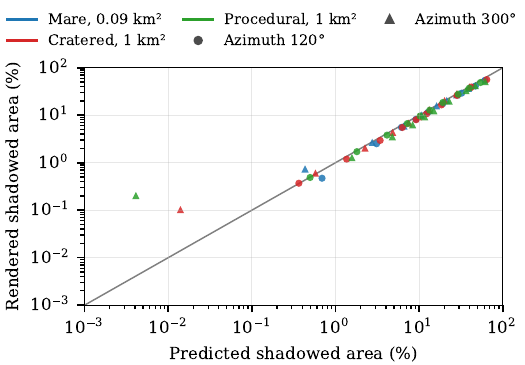}
  \caption{\small Rendered against predicted shadowed area.}
  \label{fig:shadow-area}
\end{figure}

Figure~\ref{fig:shadow-iou} illustrates the agreement between predicted and rendered shadows as a function of sun elevation. At elevations of \ang{25} and below, the mean \ac{IoU} reaches 0.79 and remains highly consistent across terrains (\numrange{0.78}{0.79}). Because shadows occupy less than \SI{2}{\percent} of the terrain at elevations above \ang{30}, minor pixel offsets disproportionately degrade \ac{IoU}. Nevertheless, the total shadowed areas correspond closely: as shown in Figure~\ref{fig:shadow-area}, the predicted and rendered areas closely track the line of perfect agreement, differing by an average of only \SI{1.5}{\percent} of the terrain area.

\subsection{Stereo Depth Under Shadows}

\begin{figure}[t]
  \centering
  \includegraphics[width=\columnwidth]{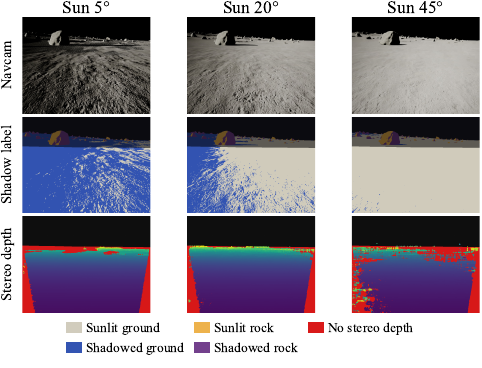}
  \caption{\small Image, shadow label, and stereo depth of the front navigation camera at three sun elevations. Red: pixels within the stereo range without a valid depth.}
  \label{fig:stereo-shadows}
\end{figure}

Because MARE inherently generates ground truth, it provides a convenient platform for evaluating perception algorithms under challenging lunar illumination. As a representative use case, we compared the stereo depth estimated by the \ac{GNC} stack against the ground-truth depth, segmenting the analysis into sunlit and shadowed pixels. The experimental setup involved the rover navigating a consistent \SI{5}{m} linear segment under six sun elevations ranging from \ang{5} to \ang{45}. With the sun positioned perpendicular to the direction of travel, this yielded between 110 and 220 frames per elevation. Evaluation was restricted to the stereo pipeline's nominal operating range of \SI{0.3}{m} to \SI{6}{m}. For both lighting classes, we quantified coverage (defined as the fraction of pixels yielding a valid stereo depth) alongside the outlier rate, which represents the fraction of covered pixels where the depth error exceeded \SI{10}{\percent} of the ground truth. As depicted in Figure~\ref{fig:stereo-shadows}, a sun elevation of \ang{5} casts most of the foreground into shadow, whereas an elevation of \ang{45} renders the terrain predominantly sunlit.

\begin{table}[t]
\centering
\setlength{\tabcolsep}{3pt}
\renewcommand{\arraystretch}{1.2}
\footnotesize
\caption{\small Stereo depth on sunlit and shadowed ground as a function of sun elevation.}
\label{tab:stereo}
\begin{tabular}{ccccccc}
\toprule
 & & & \multicolumn{2}{c}{Coverage (\%)} & \multicolumn{2}{c}{Error $>$ \SI{10}{\percent} (\%)} \\
\cmidrule(lr){4-5} \cmidrule(lr){6-7}
Sun elev. & Frames & Shadowed (\%) & Sunlit & Shadowed & Sunlit & Shadowed \\
\midrule
\ang{5}  & 189 & 86.7 & 93.4 & 84.7 & 1.2 & 0.7 \\
\ang{10} & 136 & 65.7 & 91.2 & 85.0 & 2.0 & 0.9 \\
\ang{15} & 110 & 46.7 & 91.5 & 84.2 & 1.8 & 0.7 \\
\ang{20} & 122 & 31.0 & 90.1 & 79.5 & 1.7 & 1.8 \\
\ang{30} & 114 & 9.9  & 86.1 & 57.4 & 2.7 & 9.8 \\
\ang{45} & 220 & 0.3  & 74.2 & 46.8 & 4.6 & 21.1 \\
\bottomrule
\end{tabular}
\end{table}

Table~\ref{tab:stereo} summarizes the quantitative results. The stereo pipeline consistently struggles to find matches on shadowed ground. Up to a sun elevation of \ang{20}, coverage drops from \SIrange{90}{93}{\percent} on sunlit ground to \SIrange{80}{85}{\percent} on shadowed ground. This performance gap widens at higher sun elevations, where the surviving shadows are typically small patches behind rocks and crater rims closely associated with depth discontinuities. Rocks exhibit a similar trend: \SI{99.5}{\percent} of sunlit rock pixels successfully return a valid depth, compared to only \SI{87.6}{\percent} of shadowed ones. Furthermore, sunlit ground coverage also declines as the sun rises, since higher illumination angles diminish the local shading necessary to resolve regolith texture. Notably, where the stereo system successfully computes depth, the median error remains low, staying below \SI{5}{mm} on the ground and below \SI{10}{cm} on rocks. Consequently, shadows primarily lead to missing depth data rather than inaccurate measurements, an insight made possible by the shadow masks that would otherwise be obscured by image-wide averaging.

\subsection{Dataset and Learning Baselines}

\begin{figure*}[t]
  \centering
  \includegraphics[width=\textwidth]{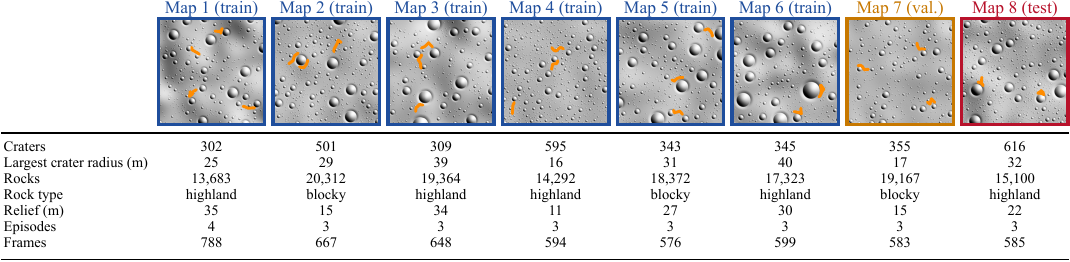}
  \caption{\small The eight dataset maps (\SI{500}{m} $\times$ \SI{500}{m}, hillshaded) with the driven routes in orange. Relief is the elevation range of each map.}
  \label{fig:dataset-maps}
\end{figure*}

\begin{figure*}[t]
  \centering
  \includegraphics[width=\textwidth]{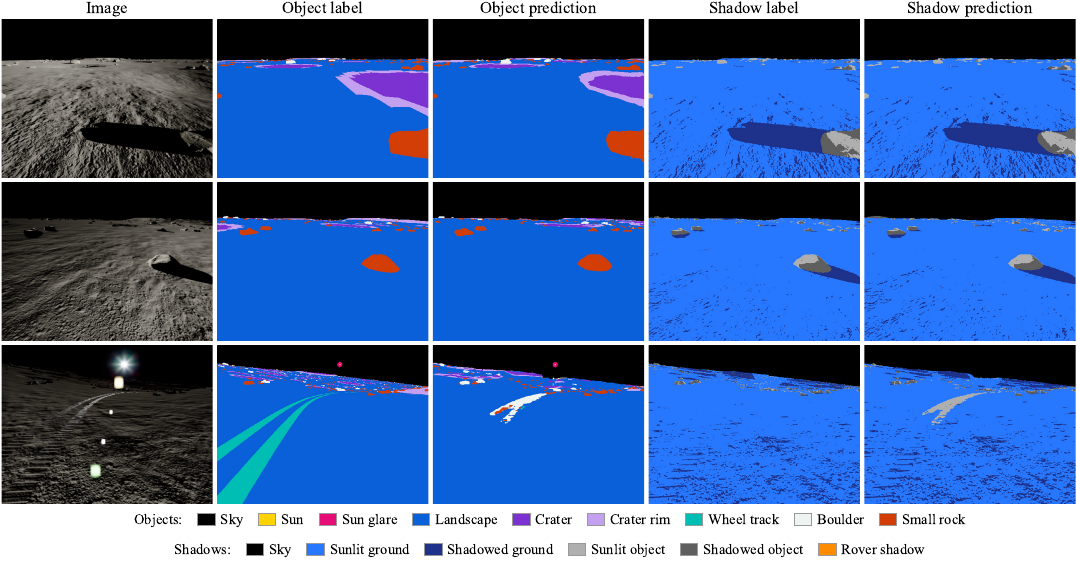}
\caption{\small Frames from the front navigation camera on the test map, shown alongside their ground-truth object and shadow labels and the corresponding U-Net predictions.}
  \label{fig:baselines}
\end{figure*}

\begin{figure*}[t]
  \centering
  \includegraphics[width=\textwidth]{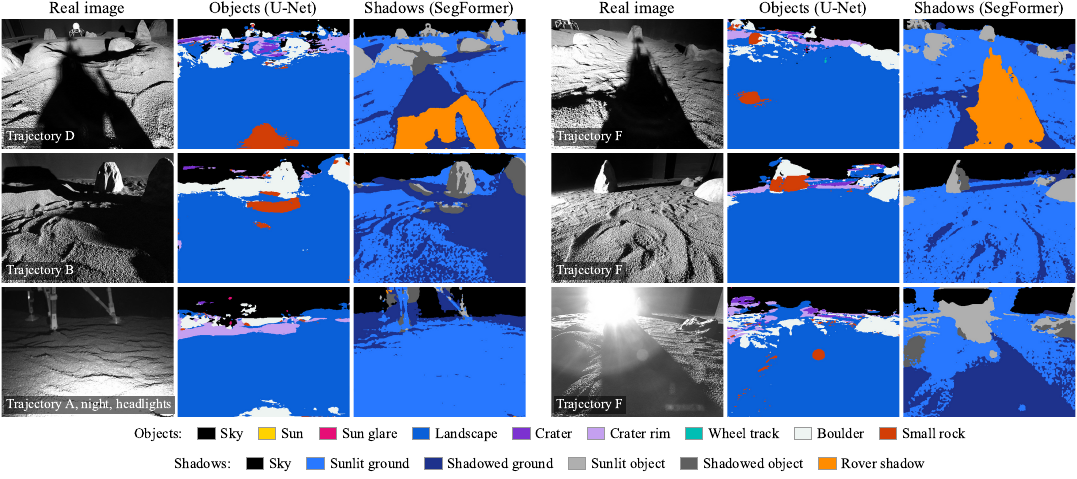}
\caption{\small Networks trained exclusively on MARE data applied to real images from the SPICE-HL3 dataset~\cite{rodriguez2026spice}, captured in the LunaLab analog facility. The figure displays object segmentation from U-Net and shadow segmentation from SegFormer-B2, representing the models that transferred best for each respective task.}
  \label{fig:baselines-real}
\end{figure*}

\begin{figure}[t]
  \centering
  \includegraphics[width=\columnwidth]{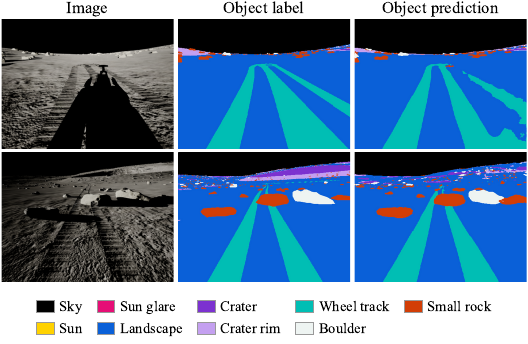}
  \caption{\small Rear camera: the wheel tracks behind the rover and the U-Net prediction.}
  \label{fig:baselines-rear}
\end{figure}

\begin{table}[t]
\centering
\setlength{\tabcolsep}{3.5pt}
\renewcommand{\arraystretch}{1.1}
\footnotesize
\caption{\small Per-class \acs{IoU} of the segmentation baselines on the test map (585 frames, sun elevations of \ang{8}, \ang{16}, and \ang{20}), with the mean \acs{IoU} over all frames and per sun elevation. Best front-camera model in bold.}
\label{tab:baselines}
\begin{tabular}{lccccc}
\toprule
 & \multicolumn{3}{c}{Front camera} & Rear & SPAD \\
\cmidrule(lr){2-4} \cmidrule(lr){5-5} \cmidrule(lr){6-6}
Class & U-Net & DeepLabV3+ & SegFormer & U-Net & U-Net \\
\midrule
\multicolumn{6}{l}{\emph{Object segmentation}} \\
Sky & \textbf{0.99} & \textbf{0.99} & \textbf{0.99} & 0.99 & 0.99 \\
Landscape & \textbf{0.90} & \textbf{0.90} & \textbf{0.90} & 0.88 & 0.88 \\
Small rock & \textbf{0.63} & 0.60 & 0.61 & 0.64 & 0.43 \\
Boulder & 0.43 & 0.39 & \textbf{0.49} & 0.54 & 0.33 \\
Crater & 0.46 & 0.46 & \textbf{0.48} & 0.25 & 0.29 \\
Crater rim & 0.21 & 0.19 & \textbf{0.23} & 0.15 & 0.17 \\
Wheel track & 0.14 & 0.15 & \textbf{0.16} & 0.94 & 0.10 \\
Sun & \textbf{0.75} & 0.65 & 0.58 & 0.74 & 0.47 \\
Sun glare & \textbf{0.84} & 0.81 & 0.82 & 0.86 & 0.69 \\
\midrule
mIoU & \textbf{0.60} & 0.57 & 0.59 & 0.67 & 0.48 \\
\quad sun 0--10$^\circ$ & \textbf{0.57} & 0.54 & 0.56 & 0.65 & 0.42 \\
\quad sun 10--20$^\circ$ & \textbf{0.68} & 0.66 & 0.66 & 0.68 & 0.57 \\
\midrule
\multicolumn{6}{l}{\emph{Shadow segmentation}} \\
Sky & \textbf{1.00} & \textbf{1.00} & \textbf{1.00} & 0.99 & 0.99 \\
Sunlit ground & \textbf{0.98} & 0.96 & 0.96 & 0.98 & 0.93 \\
Shadowed ground & \textbf{0.71} & 0.52 & 0.51 & 0.61 & 0.33 \\
Sunlit object & \textbf{0.72} & 0.67 & 0.66 & 0.76 & 0.48 \\
Shadowed object & \textbf{0.74} & 0.70 & 0.70 & 0.76 & 0.56 \\
Rover shadow & \textbf{0.99} & 0.98 & 0.98 & 0.98 & 0.97 \\
\midrule
mIoU & \textbf{0.85} & 0.80 & 0.80 & 0.85 & 0.71 \\
\quad sun 0--10$^\circ$ & \textbf{0.82} & 0.77 & 0.77 & 0.82 & 0.67 \\
\quad sun 10--20$^\circ$ & \textbf{0.87} & 0.82 & 0.82 & 0.87 & 0.73 \\
\bottomrule
\end{tabular}

\end{table}

To validate the full pipeline, we recorded a comprehensive dataset and trained baseline perception networks. Using Terrain Studio, we generated eight \SI{500}{m} $\times$ \SI{500}{m} maps varying in crater density, crater size, relief, and rock coverage (Figure~\ref{fig:dataset-maps}). Over 25 episodes, the simulated Raph Rover navigated autonomously toward three consecutive goals per episode. These goals were spaced \SIrange{10}{30}{m} apart and intentionally biased toward challenging features like craters and rocks. Illumination varied per episode, with random sun azimuths and elevations between \ang{3} and \ang{35}. Frames were extracted every \SI{0.3}{m} of travel or \SI{2}{s}, whichever occurred first, yielding 5040 frames. Each frame captures data from five cameras: a stereo pair and a central navigation camera (\num{640}~$\times$~\num{480} pixels, \ang{90} horizontal field of view, \SI{0.2}{m} stereo baseline), with the central image provided both with and without lens glare; a rear camera (\num{640}~$\times$~\num{480} pixels, \ang{90} field of view); an event camera (\num{346}~$\times$~\num{260} pixels); and a \ac{SPAD} camera (\num{256}~$\times$~\num{256} pixels). Alongside these images, every frame provides the ground-truth depth for the central camera, a lidar scan with its depth projected onto the central view, pixel-accurate object and shadow labels for the central, rear, event, and \ac{SPAD} cameras, and the exact rover pose. We partitioned the maps into six for training, one for validation, and one for testing, ensuring the test terrain remained completely unseen during training. The dataset\footnote{\url{https://huggingface.co/datasets/raulcastar/mare-lunar-rover-dataset}}, the training code\footnote{\url{https://github.com/raulcastar/MARE-Dataset-Baselines}}, and the trained models\footnote{\url{https://huggingface.co/raulcastar/mare-lunar-baselines}} are publicly available.

We trained three semantic segmentation baselines on the front camera: U-Net~\cite{ronneberger2015unet} with a ResNet-34 encoder, DeepLabV3+~\cite{chen2018deeplab} with a ResNet-50 encoder, and SegFormer-B2~\cite{xie2021segformer}. Additionally, we trained U-Net models specifically for the rear and \ac{SPAD} cameras. All encoders were pretrained on ImageNet and fine-tuned using cross-entropy and Dice losses for 30 epochs (40 epochs for the \ac{SPAD} camera), retaining the weights with the best validation score. Table~\ref{tab:baselines} details the test \ac{IoU}, while Figure~\ref{fig:baselines} illustrates predictions on the test map (real-world image transfer in Figure~\ref{fig:baselines-real} is discussed at the end of this section). Large structural classes such as sky and landscape are segmented reliably (\ac{IoU} \numrange{0.90}{0.99}). Shadows achieve a mean \ac{IoU} of 0.85, with the rover's own shadow reaching 0.99. Object segmentation performs similarly across the three architectures (mean \ac{IoU} \numrange{0.57}{0.60}). However, U-Net proves superior on shadowed ground, likely because its skip connections better preserve fine shadow boundaries. Beyond wheel tracks, crater rims remain the most challenging class (\ac{IoU} \numrange{0.19}{0.23}), as their low relief makes them visually subtle.

Sensor viewpoint heavily influences performance on specific classes, particularly wheel tracks. In the front camera, tracks only appear distantly when the rover turns or crosses its previous path, limiting the \ac{IoU} to 0.16. Conversely, the rear camera directly observes the freshly traversed terrain, boosting the track \ac{IoU} to 0.94 (Figure~\ref{fig:baselines-rear}). Sensor modality also plays a role: the \ac{SPAD} camera's lower resolution and inherent photon noise reduce its accuracy on smaller classes (mean object \ac{IoU} 0.48). While the rear camera matches the front camera in shadow segmentation (mean \ac{IoU} 0.85), the \ac{SPAD} camera drops to 0.71. Although the rover's shadow remains robustly detected (0.97), shadowed ground drops to 0.33, likely because dark regolith in shadow returns very few photons. Finally, low illumination degrades performance across all modalities (Table~\ref{tab:baselines}). When the sun elevation falls below \ang{10}, the U-Net's mean \ac{IoU} drops to 0.57 for objects, 0.82 for shadows, and 0.42 on the \ac{SPAD} camera, compared with 0.68, 0.87, and 0.57 at elevations between \ang{10} and \ang{20}.

We also evaluated monocular depth estimation, a task that predicts per-pixel distance from a single color image. Using the central navigation camera as input, we compared the predictions against ground-truth depth for pixels within \SI{0.5}{m} to \SI{80}{m}. For this, we deployed Depth Anything V2-Small~\cite{yang2024depthv2}, a network pretrained on large image collections that predicts relative (inverse) depth. In a zero-shot configuration, aligning this relative prediction to each test image via least-squares fitting yields an absolute relative error (AbsRel, defined as the mean of $|\hat{d}-d|/d$ for predicted depth $\hat{d}$ and true depth $d$) of 0.18. After fine-tuning the model for 10 epochs on the training maps to predict metric depth directly, performance improves significantly: it achieves an AbsRel of 0.027, with 98\% of pixels falling within \SI{25}{\percent} of the true depth ($\delta < 1.25$). As expected, the error scales with distance, increasing from 0.016 below \SI{5}{m} to 0.17 beyond \SI{40}{m}. We note that this high accuracy partially reflects the constant camera height and pitch of the simulated rover, which the network likely learns as a geometric prior.

Finally, to assess sim-to-real transfer, we applied the networks without adaptation to 84 real-world frames from the SPICE-HL3 dataset~\cite{rodriguez2026spice}. These images were captured by a rover in the LunaLab analog facility across seven trajectories, including one repeated under dawn, noon, and night lighting, both with and without headlights. Lacking pixel-wise labels, we evaluated the results qualitatively. Despite the domain gap introduced by grayscale imagery, distinct camera hardware, and a different physical rover, the networks successfully recover the primary scene structures, with individual architectures demonstrating distinct strengths across tasks (Figure~\ref{fig:baselines-real}). For object segmentation, U-Net effectively recognizes rocks and labels the facility floor as landscape. For shadow segmentation, SegFormer accurately isolates the rover's self-shadow and the shadows cast by rocks, whereas U-Net generates spurious rover-shadow patches. Under low illumination, the networks label the ground illuminated by the rover's headlights as sunlit terrain. Failures primarily stem from out-of-distribution appearances absent from the simulator: the facility's walls and lander mock-up are often misclassified as boulders, and the glare from the facility's artificial sun is incorrectly segmented as a sunlit object.

\subsection{Runtime}

When all sensors are actively streaming, the system publishes rover odometry and \ac{IMU} data at \SIrange{13}{28}{Hz}, navigation camera images at \SIrange{2.3}{4.3}{Hz}, and lidar point clouds at \SIrange{0.7}{1.0}{Hz}. These rates are configurable: the camera and lidar plugins set how many frames and scans they send per simulated second, and the event and \ac{SPAD} nodes set their publish rates, at the cost of more rendering for higher rates. Concurrently, the stereo depth pipeline requires \SIrange{230}{250}{ms} to complete a single cycle. Because all generated data receives a precise simulation timestamp, these varying publish rates affect the update frequency but strictly preserve temporal consistency.

\section{Conclusions and Future Work}
\label{sec:conclusions}

This paper introduced MARE, an open-source lunar simulation suite that integrates procedural terrain generation, physics-based rovers, reusable sensor plugins, pixel-accurate ground truth, and a containerized \ac{GNC} stack within a unified environment. Our validation demonstrated that shadows predicted from the terrain model closely align with the rendered scene, achieving a mean \ac{IoU} of 0.79 across three different terrains at low sun elevations. Furthermore, using the generated shadow masks showed that stereo depth estimation achieves lower pixel coverage on shadowed terrain than in sunlit areas. Networks trained on the released dataset segment shadows reliably on unseen terrain and recover the rover's shadow and rocks in real analog images without adaptation, while crater rims and wheel tracks seen from the front camera remain challenging. Future work will focus on deriving illumination directly from planetary ephemerides and systematically evaluating navigation performance across a wider range of terrains, rover configurations, and lighting conditions.

\bibliographystyle{IEEEtran}
\bibliography{references}

\vfill
\end{document}